\documentclass[letterpaper, 10pt, conference]{ieeeconf}      

\IEEEoverridecommandlockouts                              
\usepackage{graphics} 
\usepackage{epsfig} 
\usepackage{mathptmx} 
\usepackage{times} 
\usepackage{amsmath} 
\usepackage{amssymb}  
\usepackage{graphicx}
\usepackage{booktabs}
\usepackage{cite}
\usepackage{hyperref}
\usepackage[dvipsnames]{xcolor}

\title{\LARGE \bf
Pre-training on Synthetic Driving Data for Trajectory Prediction
}

\author{Yiheng Li$^{1*}$  Seth Z. Zhao$^{1*}$  Chenfeng Xu$^{1}$  Chen Tang$^{1}$  Chenran Li$^{1}$  Mingyu Ding$^{1}$ \\
Masayoshi Tomizuka$^{1}$  Wei Zhan$^{1}$ 
\thanks{$^{*}$Equal contribution}
\thanks{$^{1}$ Y. Li, S. Z. Zhao, C. Xu, C. Tang, C. Li, M. Ding, M. Tomizuka, and W. Zhan are affiliated with the University of California, Berkeley. {\tt\small \{yhli, sethzhao506, xuchenfeng, chen\_tang, chenran\_li, myding, tomizuka, wzhan\}@berkeley.edu.} Correspondence to: Chen Tang.}}

\begin{document}
\maketitle
\thispagestyle{empty}
\pagestyle{empty}

\begin{abstract}
Accumulating substantial volumes of real-world driving data proves pivotal in the realm of trajectory forecasting for autonomous driving. Given the heavy reliance of current trajectory forecasting models on data-driven methodologies, we aim to tackle the challenge of learning general trajectory forecasting representations under limited data availability. We propose a pipeline-level solution to mitigate the issue of data scarcity in trajectory forecasting. The solution is composed of two parts: firstly, we adopt HD map augmentation and trajectory synthesis for generating driving data, and then we learn representations by pre-training on them. Specifically, we apply vector transformations to reshape the maps, and then employ a rule-based model to generate trajectories on both original and augmented scenes; thus enlarging the driving data without collecting additional real ones. To foster the learning of general representations within this augmented dataset, we comprehensively explore the different pre-training strategies, including extending the concept of a Masked AutoEncoder (MAE) for trajectory forecasting. Without bells and whistles, our proposed pipeline-level solution is general, simple, yet effective: we conduct extensive experiments to demonstrate the effectiveness of our data expansion and pre-training strategies, which outperform the baseline prediction model by large margins, \emph{e.g.} 5.04\%, 3.84\% and 8.30\% in terms of $MR_6$, $minADE_6$ and $minFDE_6$. The pre-training dataset and the codes for pre-training and fine-tuning are released at \url{https://github.com/yhli123/Pretraining_on_Synthetic_Driving_Data_for_Trajectory_Prediction}.

\end{abstract}
\section{Introduction}


Trajectory forecasting is important for safely navigating autonomous vehicles in crowded traffic scenarios. The state-of-the-art trajectory forecasting models are empowered by data-driven supervised learning approaches, whose performance heavily relies on the scale of motion data available for training~\cite{Waymo, yu2020bdd100k, PreTram}. However, driving data is expensive and time-consuming to collect and annotate, which hinders cost-efficient scaling of training data as in Natural Language Processing (NLP) and Computer Vision (CV). Data-collection vehicles with sophisticated sensors need to run on public roads to collect traffic data. For instance, the Argoverse V1.1 dataset~\cite{argo19} and the Waymo Motion Dataset~\cite{Waymo} consist of 324K and 103K driving scenes, which add up to a total of 320 and 574 hours of driving data, respectively. Notably, a substantially larger amount of raw data needs to be collected to filter out the given amount of high-quality training data. In addition to data collection, a significant amount of human labor is required to annotate the raw data into a cohesive dataset\textemdash traffic participants need to be annotated, and HD maps need to be aligned and integrated with the traffic data. 

\begin{table}
  \caption{Effortless Driving Data Generation.}
  \label{tab:effort}
  \centering
  \setlength{\tabcolsep}{10pt}
  \vspace{-6pt}
  \begin{tabular}{l|cc}
    \toprule[1.0pt]
    Dataset & Driving Hours & Scenes\\
    \midrule
    Argoverse v1.1 & 320 & 324k \\
    
    Synthetic Dataset & +0 & +370k \\
  \bottomrule[1.0pt]
\end{tabular}
\vspace{-6pt}
\end{table}

\begin{figure}
       \centering
       \includegraphics[width=0.75\linewidth]{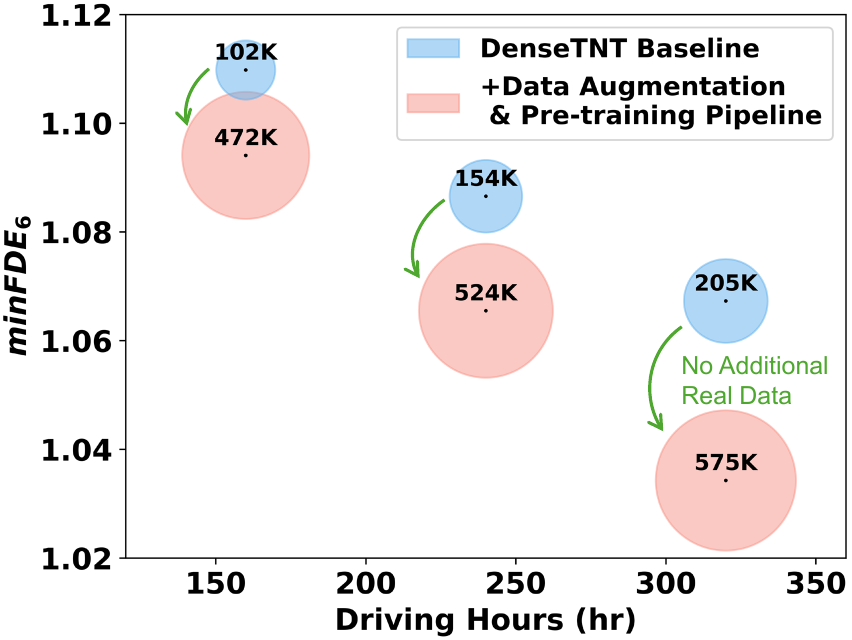}
       \vspace{-6pt}
   \caption{\textbf{Our data synthesis and self-supervised pre-training pipeline enhances prediction performance without extra real-world driving data.} Each circle's area is proportional to the total number of synthetic and real driving scenes used, as indicated within. (A lower $minFDE_6$ is preferable.)}
   \vspace{-6pt}
   \label{fig:efficiency}
\end{figure}

In this work, we sought to \emph{synthetic data} as a solution to break through the data bottleneck in trajectory forecasting. Compared to collecting and processing real-world data, which involves huge human labor, automatically generating synthetic data is effortless. As shown in Tab.~\ref{tab:effort}, our proposed data generation method only takes 20 \textit{computational hours} to generate a comparable number of scenes to the real-world dataset, which requires hundreds of \textit{human driving hours}. More importantly, as shown in Fig.~\ref{fig:efficiency}, the synthetic data can significantly improve the prediction accuracy and data efficiency when paired with our proposed training pipeline.

Concretely, we propose a pipeline-level solution with two parts: we generate the synthetic data and then learn representation by pre-training on them. The generation part consists of a map augmentation and a trajectory generation process. For map augmentation, we adopt the vector-transformation method~\cite{bahari2022sattack} to convert the linear lanes found in real-world maps into curved lanes within a defined range of sharpness and angle, as shown in Fig.~\ref{fig:pipeline} (upper-left). The augmented curved roads are introduced to diversify the map data. Subsequently, we navigate a simulated vehicle on both real-world and augmented maps with a rule-based planner~\cite{Yin2021IterativeIP} to generate trajectory data. The rule-based planner leverages prior knowledge to facilitate realistic generated trajectories. 

Leveraging the generated data is an open question in robotics, NLP, and computer vision \cite{wang2024pretraining,he2023synthetic1,he2023synthetic2, tian2024what}. We conduct an extensive study on various training algorithms that use synthetic data to learn general representations for trajectory forecasting. In particular, we compare three different paradigms: augmenting the real-world prediction dataset~\cite{zhong2020random, Ryu_2023_CVPR_Domain_Aug, Kim_2021_ICCV_Point_Cloud_Aug}, supervised pre-training~\cite{He_2019_ICCV_ImagenetPT,WeakSupervised_2018_ECCV,pmlr-L2T}, and self-supervised pre-training~\cite{vincent2010stackeddenoised,i-jepa,simclr, he2020momentum, MaskedAutoencoders2021, BERT, Huang_2021_ICCV,yao_2022_pevl, chen_2022_pix2seq, CLIP, zhang2023metatransformer}. The goal is to train a generalizable scene representation on the synthetic data that oversees detailed domain-specific information and can benefit cross-domain fine-tuning. Extensive experiments on the Argoverse dataset with denseTNT backbone demonstrate that self-supervised pre-training outperforms the other two methods. With our synthetic data generation and self-supervised pre-training pipeline, the fine-tuned model outperforms the baseline model by a large margin\textemdash 5.04\%, 3.84\%, and 8.30\% in terms of $MR_6$, $minADE_6$ and $minFDE_6$. Detailed ablation studies are then conducted to shed light on the influence of some subtle factors on prediction performance, such as information leaks and map augmentation. 

We summarize our contributions below:
\begin{itemize}
    \item We propose a simple yet effective pipeline for learning general representations in trajectory forecasting at the low-data regime. Our pipeline highlights the synergy of the synthetic map and trajectory generation and representation learning via pre-training on the synthetic data. This pipeline does not introduce any extra human effort while drastically improving the volume of data for the training and enhancing the representation of learning under the scarcity of real driving data.

    \item We conduct extensive experiments to pinpoint the optimal training scheme to utilize the synthetic data. Our results show that self-supervised pre-training performs better than the other widely explored alternatives, such as directly augmenting the training dataset and supervised pre-training.

    \item  With our synthetic data generation and self-supervised pre-training pipeline, the fine-tuned model outperforms the baseline model by a large margin\textemdash 5.04\%, 3.84\%, and 8.30\% in terms of $MR_6$, $minADE_6$ and $minFDE_6$. 
\end{itemize}

\section{Related Works}
\subsection{Data Augmentation in Trajectory Prediction}
Data scarcity is a bottleneck for exploring the model's capacity for better performance \cite{PreTram}. To address this issue, numerous data augmentation techniques, such as the application of image transformations \cite{simclr, dataaug, xiang2018posecnn} and utilization of synthetic data \cite{he2022fs6d, copypaste, dttd, xiang2018posecnn}, are introduced in the CV community to augment or replicate seldom-seen real-world scenes. In the trajectory prediction problem, the unique properties of trajectory and map data necessitate a specific design for data augmentation or generation that is both reasonable and cost-effective. In \cite{huaweipretrain}'s approach, a simple augmentation technique is applied to the vehicle's velocity properties to mimic unpredictable driver behaviors in anticipation of future trajectories based on past trajectories. Additionally, \cite{bahari2022sattack} proposes using map augmentation techniques to alter the topological semantics of map information as adversarial attacks on current trajectory prediction models. While these methods expand the dataset size, their additional trajectories are simple geometric transformations of existing lane elements or trajectories without carefully considering the transformed trajectories' feasibility and realism. In our work, we enhance the map data to supplement limited features, like curves and turns, and employ a rule-based simulation to mimic reasonable traffic behaviors for trajectory generation.

\begin{figure*}[t]
        \centering
        \includegraphics[width=1.00\linewidth]{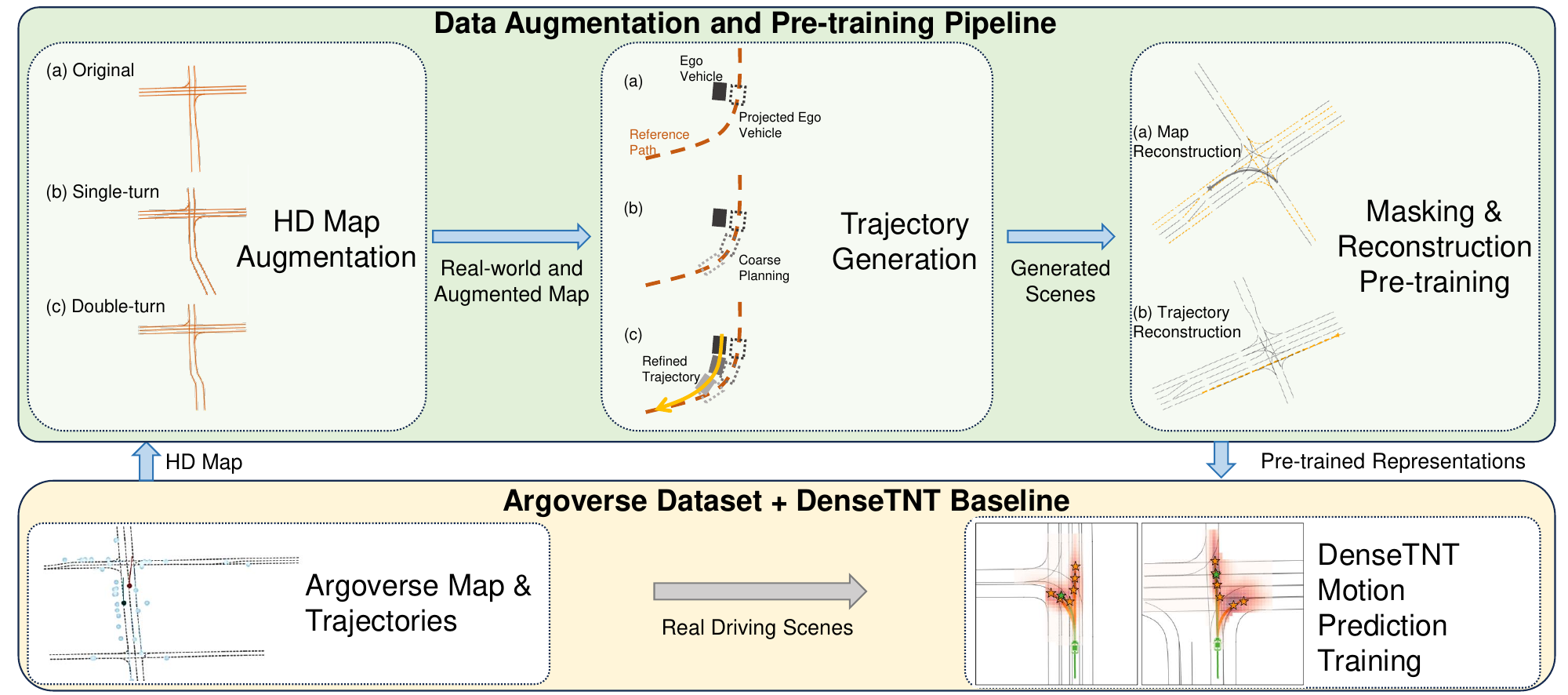}
        \vspace{-6pt}
    \caption{\textbf{Pipeline of driving data synthesis and utilization.} We augment the map and generate trajectory on it to acquire synthetic motion data, which are then used for pre-training via masking and reconstruction. The pre-trained model is used to initialize the backbone model for fine-tuning.}
    \label{fig:pipeline}
    \vspace{-6pt}
\end{figure*}
\subsection{Pre-training in Trajectory Prediction} 

 Pre-training is effective for the data shortage problem in trajectory prediction. In \cite{PreTram}, contrastive learning was used to cluster similar map patches and establish associations between map patches and the trajectory, thereby fully utilizing the existing map data. In \cite{ssl-lanes}, four pre-training tasks were initiated to enhance the encoding, including randomly masking and recovering parts of the features. Concurrently, \cite{trajMAE, cheng2023forecastmae} used MAE to pre-train the encoder. However, these methods did not introduce any new data. Another work \cite{huaweipretrain} generated a pseudo trajectory that strictly follows the lanes for pre-training. However, this strict compliance hinders the realism of the pseudo trajectories. In contrast, our work aims to generate realistic synthetic trajectories to mitigate the domain gap between the synthetic and real data. 
 
 While limited efforts on pre-training exist in the trajectory forecasting literature, some auxiliary tasks have been introduced to regularize trajectory prediction models during training. \cite{2020vectornet} introduced a feature-masking task to encourage feature interactions, which inspired us to use methods in CV \cite{MaskedAutoencoders2021} and NLP \cite{BERT} to stimulate feature interactions in the learned scene representation. Taking cues from these works, we investigated masked reconstruction as a self-supervised pre-training method to utilize the synthetic data. 

\section{A pipeline-level solution of learning general representations} 

\subsection{Data Generation} \label{Synthetic}
Our data synthesis method is simple yet enables expanding the diversity of trajectory data. The key components include a map augmentation module and a model-based planning model to generate trajectories.

\subsubsection{Map Data Augmentation}
Considering the lack of curve trajectory in the real-world training dataset, we add the map augmentation module to our data synthesis pipeline. Inspired by the conditional adversarial scene generation process proposed in \cite{bahari2022sattack}, we design transformation functions that alter the original map topology, as shown in Fig. \ref{fig:pipeline} (upper-left). Given that each scene is composed of a set of scene points, we define the transformation on each scene point $(s_x, s_y)$ in the following form:
\begin{equation}
\overline{s} = (s_x, s_y + f(s_x - b))
\end{equation}
where $\overline{s}$ is the transformed point, $f$ is the single-variable transformation function, and $b$ is a parameter that determines the region of applying the transformation, i.e., we only modify the scene points whose x-coordinates surpass $b$. We further adopt the following two transformation functions for our map augmentation process.

\textbf{Single-turn} introduces a single turning on the roadway. The transformation function for single-turn is defined as:
\begin{equation}
f_{\mathrm{single-turn}}(s_x) = 
    \begin{cases}
  0, & s_x < 0, \\
  q_{\alpha}(s_x) , & 0 \leq s_x \leq s_t \\
  (s_x - s_t)q_{\alpha}'(s_t) + q_{\alpha}(s_t), & s_t < s_x \\
    \end{cases}
\end{equation}
where $s_t$ represents the length of the turn and $q_\alpha$ is an auxiliary function defined by 
\begin{equation}
q_\alpha(s_x) = \frac{\alpha_1}{s_t^{\alpha_2}} s_x^{\alpha_2}
\end{equation}
where $\alpha_1, \alpha_2$ are the parameters to control the turn's sharpness and angle. Note the $q_{\alpha}'$ is simply the derivative of $q_{\alpha}$. Our formulation of $f_{\mathrm{single-turn}}(s_x)$ is continuously differentiable and thus makes a smooth augmented single-turn curve.

\textbf{Double-turn} introduces two consecutive smooth single-turns in opposite directions to the road. The transformation function for double-turn is based on the that of single-turn:
\begin{equation}
f_{\mathrm{double-turn}}(s_x) = f_{\mathrm{single-turn}}(s_x) - f_{\mathrm{single-turn}}(s_x - \beta)
\end{equation}
where $\beta$ is the distance between two turns. In practice, we randomly select single-turn or double-turn augmented lanes and randomly assign values in a specific range to ensure a reasonable extent of transformation, as shown in Tab.~\ref{tab:param_traj}. 

\begin{table}[h]
\caption{Parameter selection for data synthesis.}
  \label{tab:param_traj}
  \centering
  \setlength{\tabcolsep}{14pt}
  \resizebox{\linewidth}{!}
  {
  \begin{tabular}{c|c|c|c}
    \toprule[1.0pt]
    Traj. Param. & Value & Map Param. & Value \\
    \midrule
    $a$ & $\in [-2, 1]$ & $s_x$ & 10  \\
    $w_1$ & 5 & $\alpha_1$ & $\in [1, 10]$ \\
    $w_2$ & 5 & $\alpha_2$ & 20   \\
    $w_3$ & 1 & $s_t$ & 10  \\
    $v_d$ & $\in [6, 15]$ & $\beta$ & 20 \\ 
  \bottomrule[1.0pt]
\end{tabular}
}
\end{table}

\subsubsection{Trajectory Data Generation}
To generate labeled trajectory data on both real-world and augmented maps, we implement a trajectory data generator to synthesize pseudo-expert trajectories in single-car scenarios, as illustrated in Fig. \ref{fig:pipeline} (upper-middle). Following the motion data generation pipeline proposed in \cite{Yin2021IterativeIP}, we project its position onto the nearest reference path obtained from the training dataset to determine the optimal trajectory for the ego vehicle. Subsequently, we utilize the A$^*$ planning method on this reference path to generate a preliminary, coarse trajectory. Specifically, we define a node $n_i$ in the search space by a tuple $(s_i, v_i, t_i)$, which respectively represents the ego vehicle's coordinate, velocity, and time on the reference path. A transition from $n_i$ to $n_{i+1}$ defines the $n_{i+1}$ by:
\begin{equation}
(s_{i+1}, v_{i+1}, t_{i+1}) = (s_i + v_i\delta t + \frac{1}{2}a\delta t^2, v_i + a\delta t, t_i + \delta t)
\end{equation}
where $a$ denotes the acceleration and $\delta t$ denotes the minimum time interval for coarse planning. The cost function of the transition is given as:
\begin{equation}
\mathcal{C}(n_i, a, n_{i+1}) = w_1a^2 + w_2\kappa(s_{i+1})v_{i+1}^2 + w_3(v_{i+1} - v_d)^2
\end{equation}
where $w_1, w_2, w_3$ are weights for each term, $\kappa(s_{i+1})$ denotes curvature of reference path at $s_{i+1}$, and $v_d$ denotes desired driving velocity set in prior. The planning process concludes when the time associated with the searched node surpasses the end time $t_g$, resulting in the conversion of the resultant path into a coarse trajectory $s_{c}(t)$ within the global frame. The hyperparameters adopted for the synthetic trajectory generation pipeline are listed in Table \ref{tab:param_traj}. Finally, we introduce an additional refinement procedure to map the final pseudo-expert trajectory's initial state to the ego car's current state. Concretely, we solve the following optimization problem to acquire the refined trajectory $s_{r}(t)$:
\begin{align}
\footnotesize
    \min_{s_{r}} & \sum^{t_g}_{T=\delta\hat{t}} \omega_1 a_{r}(T)^2 + \omega_2 j_{r}(T)^2 
    + \sum^{t_g}_{T=k\delta \hat{t}} \omega_3(s_{r}(T) - s_{c}(T))^2 \nonumber \\
    \text{s.t.} & \quad  v_{r}(T_0) = v_0\quad \text{and} \quad s_{r}(T_0) = s_0
\end{align}
where $\omega_1, \omega_2, \omega_3$ are weights for each term, $a_{r}, j_{r}$ are refined acceleration and jerk, $\delta \hat{t}$ is the fine-grained time step such that $k \delta \hat{t} = \delta t$.  
$v_0, s_0$ corresponds to the ego car's initial velocity and position.



\subsection{Representation Learning on Synthesis Data}\label{pt_methods}
Representation learning on synthesis data is widely explored in other fields \cite{he2023synthetic1,he2023synthetic2,wang2024pretraining}. As far as we know, we are the first to make use of generated data in trajectory forecasting for autonomous driving. We comprehensively study several commonly used approaches to identify the best strategy. The first approach is to directly combine the synthetic and real-world data for training, i.e., treating them identically. While this method is straightforward, it introduces bias into the model due to the domain gap between the synthetic and real-world data. Another method is supervised pre-training: we first pre-train the model for the supervised trajectory prediction task on the synthetic dataset and then fine-tune the model on the real-world dataset. While the pre-training and fine-tuning tasks are both supervised prediction tasks, the two-stage scheme is introduced to mitigate the bias caused by the domain gap. Then, we investigate the self-supervised pre-training method elaborated as follows:

\subsubsection{Self-supervised Pre-training Framework}\label{mae-pretrain}
We adapt MAE~\cite{MaskedAutoencoders2021} into a self-supervised pre-training framework for trajectory forecasting. The goal is to learn a generalizable scene representation that can capture the interactions between lanes and the ones between lanes and trajectories from the synthetic data well. During the pre-training stage, a specific percentage of trajectories or map lanes are masked, challenging the model to reconstruct the masked segments. The masked objects are replaced with mask tokens, which are learned and shared parameters that signal the existence of a masked object to be reconstructed~\cite{MaskedAutoencoders2021}. Positional encoding is added to provide the subordinate and sequential information. Though specific information is removed in mask tokens, they still provide the model with where the masked information is. Such information leak will lead the model to find a shortcut to interpolate missing information instead of reconstructing them by context understanding~\cite{convmae}. Only the unmasked elements are processed through the backbone's encoder to prevent this. After that, the masked tokens, in combination with the encoded unmasked parts, are fed into a shallow transformer decoder to reconstruct the masked parts. Upon completion of the pre-training phase, the backbone model is initialized using the encoder parameters of the pre-trained model and then fine-tuned on real-world data for the prediction task. This approach enables the model to acquire general semantics during pre-training while discarding extraneous details. 

\subsubsection{Masking Strategy}
As each lane or trajectory is not as strongly correlated to other lanes or trajectories as the pixels in a figure, we need to design masking strategies to apply MAE to the trajectory prediction problem. Specifically, we propose and compare different design options for masking and reconstruction during the pre-training stage:

\textbf{Map reconstruction} is proposed to foster the encoder to capture the lane features and their interactions. The masking policies and reconstruction procedures are illustrated in Fig. \ref{fig:pipeline} (upper-right). We randomly choose 50\% lanes as masking lanes, replacing everything but the coordinate information for their first points with mask tokens. These masked objects, together with the encoded features of the unmasked objects, are then fed into the decoder. Utilizing lane-wise interactions, the feature of the masked lanes would be recovered. Point-wise L1 loss is employed across all our experiments. The loss function of map reconstruction $l_{map}$ is the difference between the reconstructed lane and the ground-truth one.

\textbf{Trajectory reconstruction} is introduced to improve the model's understanding of the trajectory features and the interactions between the trajectory and map elements, as shown in Fig. \ref{fig:pipeline} (upper-right). Since the synthetic scene only consists of one agent, only a single trajectory is masked in each scene. We retain its starting point in the masked object as in map reconstruction. To prevent mode collapse, the decoder is allowed to produce six potential trajectories. We select the one with the minimum loss to define the loss function $l_{traj}$. Additionally, to ensure that all modes are activated, we add the weighted reconstruction errors of the remaining modes to the loss function as regularization. A weight of 0.05 is used in our experiments. 

\textbf{Combination of Map and Trajectory Reconstructions} integrates both map and trajectory reconstruction tasks by randomly selecting a percentage of scenes within a mini-batch to conduct either of the reconstructions. This selection is achieved by generating a random number prior to masking each scene. As a result, the same scene could undergo different reconstruction tasks across various epochs. The overall loss function is computed as the average of the individual scene-specific losses previously described.


\section{Experiments}

\begin{figure*}
        \centering
        \includegraphics[width=0.70\linewidth]{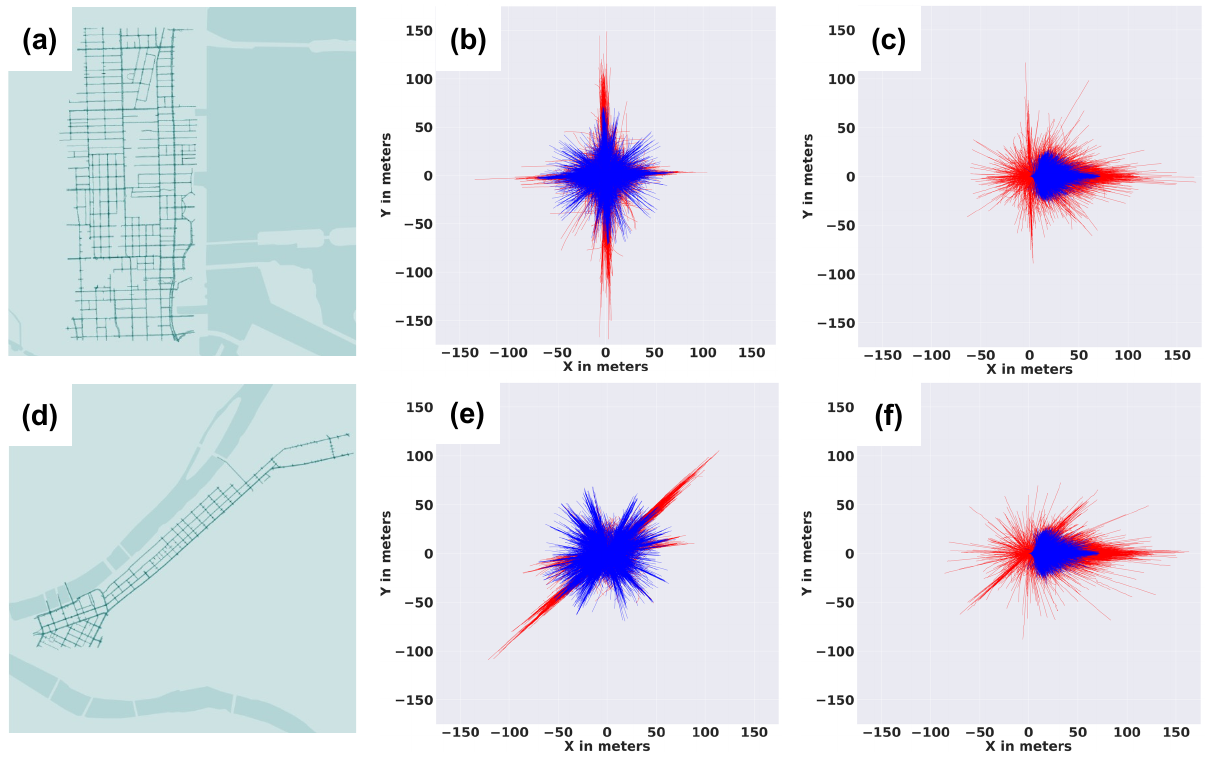}
        \vspace{-6pt}
    \caption{\textbf{Data distribution comparison between synthetic dataset {\color{blue}(blue)} and the real-world dataset {\color{red}(red)} in terms of the speed and direction properties.} (a) represents the map representation for MIA city. (b) represents the trajectory distribution of scenes in (a), showing a pattern of divergence in velocity properties. (c) represents the trajectory distribution of scenes in (b) rotated to the same initial direction, demonstrating a pattern of divergence in direction properties. (d)(e)(f) are the counterparts of (a)(b)(c) in PIT city.}
    \label{fig:rotation_vis}
    \vspace{-6pt}
\end{figure*}

\subsection{Synthetic Dataset} 

Our synthetic dataset is generated in accordance with the methods outlined in Sec. \ref{Synthetic}. The map is sourced from the HD maps in the Argoverse v1.1 Motion Forecasting Dataset \cite{argo19}. We generate trajectories on both the original map and the augmented map. The resultant generated dataset contains 370k driving scenes, where 205k are generated on the original map and 165k are generated on the augmented map. Each generated scene contains the vehicle's position spanning 5 seconds, following the format of \cite{argo19}. As visualized in Fig.~\ref{fig:rotation_vis}, the generated data align well with a vast spectrum of velocities in real-world data distribution. While our synthetic dataset encompasses a broader spectrum of velocity than the real-world data, it is still important to note that vehicles may still drive under extreme velocity in some long-tailed events present in the real-world data. Additionally, the real-world dataset encompasses a multitude of scenarios with pronounced turning angles, such as U-turns, which pose challenges in accurate generation within the synthetic dataset. The discrepancy in data distributions validates that self-supervised pre-training is necessary to mitigate the impact of distributional shifts.

\subsection{Prediction Experiment Setup}
\textbf{Argoverse v1.1 Motion Forecasting Dataset.}
The Argoverse v1.1 Motion Forecasting Dataset \cite{argo19} is a renowned dataset frequently employed in vehicle motion prediction studies. It consists of 324k scenes, each spanning 5 seconds and sampled at 10 Hz. In every scene, the initial 2 seconds serve as the history, while the subsequent 3 seconds are designated for prediction. The dataset also includes HD maps collected from Pittsburgh and Miami, covering all the scenes in the dataset. There are 205k and 39k scenes in the training and validation sets, respectively, in which the validation set is used for evaluating our model.

\textbf{Evaluation Metrics.} 
Following Argoverse's benchmark, we use miss rate ($MR_6$), minimum final displacement error ($minFDE_6$), and minimum average displacement error ($minADE_6$) as the evaluation metrics. The subscripts mean that the minima are computed over six predicted trajectories.

\textbf{Implementation Details.} 
In our experiments, we adopt DenseTNT \cite{densetnt} as the prediction backbone. DenseTNT uses Vectornet \cite{2020vectornet} as its encoder, which adopts a vectorized scene representation, i.e., the lanes and trajectories are represented as vectors, which makes the masking process straightforward. The pre-training is conducted on an NVIDIA A6000 GPU with a batch size of 256. The learning rate is $10^{-5}$ for trajectory reconstruction and $10^{-3}$ for map reconstruction. We use a learning rate of $10^{-4}$ for mixed map and trajectory reconstructions. Except for the GPU, all other fine-tuning settings are the same as in DenseTNT \cite{densetnt}. During evaluation, we choose the 100ms optimization for $minFDE_6$.

\subsection{Performance of the Proposed Pipeline}

\begin{table}
  \caption{Performance of Self-Supervised Pre-Training.}
  \label{tab:bestresult}
  \centering
  \vspace{-6pt}
  \resizebox{\linewidth}{!}
  {
  \begin{tabular}{l|ccc}
    \toprule[1.0pt]
    
    Method & $MR_6$(\%) ($\downarrow$) & $minFDE_6$ ($\downarrow$) & $minADE_6$ ($\downarrow$)\\
    \midrule
    Baseline & 9.73 & 1.0673 & 0.8052\\
    \midrule
    Map Reconstruction & \textbf{9.20 (-5.45\%)} & 1.0343 (-3.09\%) & 0.7571 (-5.97\%)\\
    Trajectory Reconstruction & 9.24 (-5.04\%) & \textbf{1.0263 (-3.84\%)} & 0.7384 (-8.30\%)\\
    Combined Reconstruction & 9.27 (-4.73\%) & 1.0349 (-3.04\%) & \textbf{0.7284 (-9.54\%)}\\
    \bottomrule[1.2pt]
    \end{tabular}
    }
    \vspace{-4pt}
\end{table}

We present the trajectory prediction performances of our pipeline with different self-supervised pre-training tasks in Tab. \ref{tab:bestresult}. The choice of adopting self-supervised pre-training for generated data results from thorough performance comparisons of different representation learning methods in Part \ref{pt_comparison}. The experiment results show that all of the self-supervised pre-training tasks can substantially improve the prediction accuracy compared with the baseline. Specifically, with map reconstruction, an $MR_6$ of 9.20\% is observed, which is 5.04\% lower than the baseline. Trajectory reconstruction pre-training helps improve the $minFDE_6$ by 3.84\% compared to the baseline. Combining them together, when we allocate 70\% scenes in a batch to map reconstruction and 30\% scenes to trajectory reconstruction, we achieve a $minADE_6$ of 0.7284, marking an improvement of 9.54\% against the baseline.

\begin{figure*}[t]
        \centering
        \includegraphics[width=0.95\linewidth]{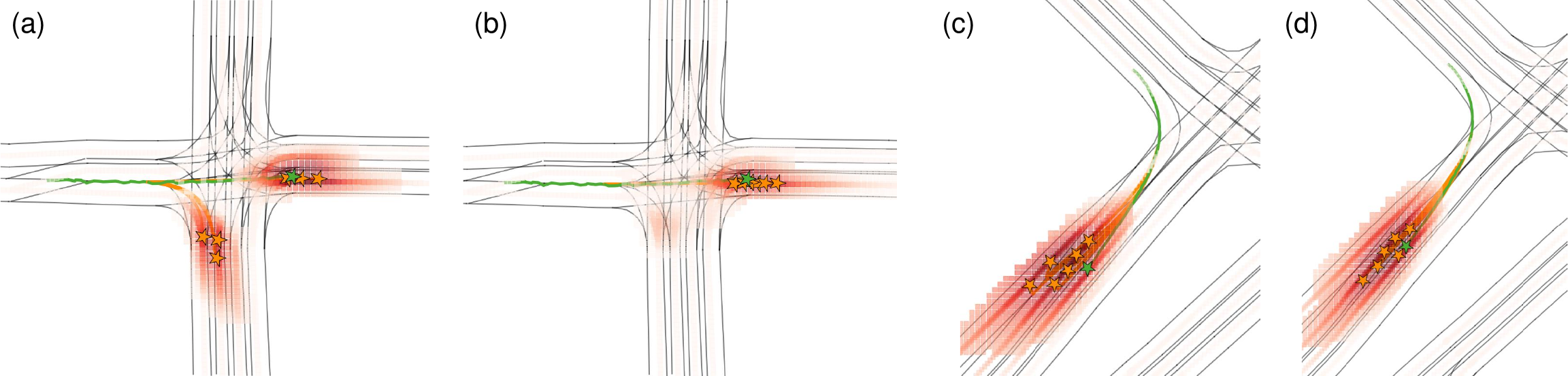}
        \vspace{-6pt}
\caption{\textbf{Performance comparison without or with pre-training.} The \textcolor{gray}{gray} lines indicate lane boundaries. The \textcolor{green}{green} line and star indicate the true trajectory and its last point, while \textcolor{orange}{orange} ones are the predicted ones. The \textcolor{orange}{orange} background shows the possibility of each point being the predicted last point of the trajectory. (a) and (b) show the prediction results without or with trajectory pre-training. (c) and (d) illustrate the performance without or with map pre-training.}
    \label{fig:compare}
    \vspace{-6pt}
\end{figure*}

To elucidate the impact of our synthetic data and the pre-training, we provide qualitative examples in Fig. \ref{fig:compare}. Fig.~\ref{fig:compare} (a) and (b) contrast prediction outcomes without and with trajectory reconstruction pre-training. We observe that the model without pre-training predicts potential right-turn trajectories. However, since the target vehicle is not in the right-turn lane, its intention to go straight should be able to be identified, if the encoder captures the spatial relation between the historical trajectory and the lanes. In contrast, trajectory pre-training strengthens the model's confidence in discerning the go-straight intention. Sub-figures (c) and (d) compare the prediction outcomes of the models without and with map reconstruction pre-training. In these sub-figures, the right-turn lane eventually merges with the 2$^\mathrm{nd}$ lane from the right. We observe that with map pre-training, the model can more accurately predict that the vehicle will not merge into the 1st lane from the right. This result aligns with expectations\textemdash map reconstruction enhances the model's understanding of lane connectivity, thereby improving prediction accuracy.

\section{Discussion}
In this section, we elaborate on some detailed design choices we made. In Sec ~\ref{sec:data} and Sec.~\ref{sec:map}, we validate the necessity of driving data synthesis and map augmentation. In Sec ~\ref{pt_comparison}, we compare different representation learning methods for utilizing generated data. In Sec.~\ref{sec:info}, we look into the information leak problem and validate our pre-training model architecture choice, which avoids information leaks and effectively improves the fine-tuned model's performance. 

\subsection{Benefits of Synthetic Driving Data}\label{sec:data}

\begin{table}
  \caption{Contributions of Synthetic Data.}
  \label{tab:datacompare}
  \centering
  \setlength{\tabcolsep}{8pt}
  \vspace{-6pt}
  \begin{tabular}{l|ccc}
    \toprule[1.2pt]
    Pretrain Dataset & $MR_6$(\%) & $minFDE_6$ & $minADE_6$\\
    \midrule
    Baseline & 9.73 & 1.0673 & 0.8052\\
    \midrule
    Argoverse Dataset & 9.68 & 1.0517 & 0.7571\\
    Synthetic Dataset & \textbf{9.24} & \textbf{1.0263} & \textbf{0.7384}\\
  \bottomrule[1.2pt]
\end{tabular}
\vspace{-6pt}
\end{table}

A straightforward baseline is directly performing pre-training on the given real dataset. To demonstrate the superiority of using the effortlessly obtained synthetic data, we compare the performance of the trajectory forecasting model pre-trained on the synthetic and real-world data, respectively, as shown in Tab. \ref{tab:datacompare}. We observe that pre-training on the real-world dataset does not bring as much improvement as utilizing our synthetic data. Specifically, pre-training on the synthetic data improves that on the real data over 0.44\% points, 0.0254, and 0.0187 points in terms of $MR_6$, $minFDE_6$, and $minADE_6$, respectively, which emphasizes the benefits of our method in both its effectiveness and its efficiency in exploiting real-world driving data. 

\subsection{Benefits of Map Augmentation}\label{sec:map}

\begin{table}
  \caption{Contributions of map augmentation.}
  \label{tab:augcompare}
  \centering
  \setlength{\tabcolsep}{8pt}
  \vspace{-6pt}
  \begin{tabular}{l|ccc}
    \toprule[1.2pt]
    Pretrain Dataset & $MR_6$(\%) & $minFDE_6$ & $minADE_6$\\
    \midrule
    Baseline & 9.73 & 1.0673 & 0.8052\\
    \midrule
    Original Map & 9.40 & 1.0517 & 0.7694\\
    Original+Augmented Map & \textbf{9.20} & \textbf{1.0343} & \textbf{0.7571}\\
  \bottomrule[1.2pt]
\end{tabular}
\vspace{-6pt}
\end{table}

Previous work \cite{PreTram} shows that directly pre-training for learning representations regarding the HD map also matters, and recent work \cite{hallgarten2024stay} also indicates the importance of adapting trajectory prediction methods to diverse maps. Therefore, following \cite{PreTram}, we assess the performance of models pre-trained with and without scenes generated on the augmented maps. We also utilize map reconstruction as the proxy task for pre-training. The comparison experiments are shown in Tab. \ref{tab:augcompare}.  It can be seen that the model trained on data with augmented maps significantly outperforms the other. This underscores the advantage of map augmentation in improving prediction models.

\subsection{Comparison of Different Representation Learning Methods}\label{pt_comparison}

\begin{table}
  \caption{Method Comparison for Using Synthetic Data. PT=Pre-training}
  \label{tab:methodcompare}
  \centering
  \vspace{-6pt}
  \resizebox{\linewidth}{!}{\begin{tabular}{l|c|ccc}
    \toprule[1.2pt]
    
    Method & Pretrain Task & $MR_6$(\%) ($\downarrow$) & $minFDE_6$ ($\downarrow$) & $minADE_6$ ($\downarrow$)\\
    \midrule
    Baseline & / & 9.73 & 1.0673 & 0.8052\\
    \midrule
    Dataset Augmentation & / & 9.45 & 1.0567 & 0.7432 \\
    Supervised whole PT & Prediction & 9.34 & 1.0371 & 0.7551 \\
    Supervised encoder PT & Prediction & \textbf{9.20} & 1.0369 & 0.7639 \\
    Self-supervised PT (Ours) & Reconstruction & 9.24 & \textbf{1.0263} & \textbf{0.7384}\\
  \bottomrule[1.2pt]
\end{tabular}}
\vspace{-6pt}
\end{table}

We compare the performances of multiple methods described in Sec.~\ref{pt_methods}. The results are shown in Tab. \ref{tab:methodcompare}. While directly augmenting the training dataset can slightly improve the prediction accuracy, it does not perform as well as pre-training approaches. The result is consistent with our motivation behind adopting pre-training methods. We compared two alternative fine-tuning strategies for supervised pre-training methods: initializing the entire model with the pre-trained model or only initializing the \emph{encoder} with the pre-trained one. We found that the decoder does not benefit from loading the parameters from the pre-trained model. We hypothesize that the domain-specific nuances learned by the decoder during supervised pre-training might not generalize well to real-world scenarios. This implies that synthetic data is most suitable for specifically improving representation learning rather than directly improving the entire backbone model for trajectory prediction. To this end, self-supervised pre-training should be the most efficient way to utilize the synthetic data, as the training tasks are deliberately designed for learning generalizable scene representation, which is supported by the experimental results\textemdash The self-supervised pre-trained model achieves the best prediction accuracy after fine-tuning; therefore it is chosen in our pipeline.

\subsection{Information Leak in Self-supervised Pre-training}\label{sec:info}

\begin{table}
  \caption{Influence of Information Leak.
  PT=pre-training.}
  \label{tab:unsupcompare}
  \centering
  \setlength{\tabcolsep}{9pt}
  \vspace{-6pt}
  \begin{tabular}{l|ccc}
    \toprule[1.2pt]
    Pretrain Method & $MR_6$(\%) & $minFDE_6$ & $minADE_6$\\
    \midrule
    Baseline & 9.73 & 1.0673 & 0.8052\\
    \midrule
    No Info Leak Traj PT & 9.24 & \textbf{1.0263} & \textbf{0.7384}\\
    W/ Info Leak Traj PT & 9.45 & 1.0520 & 0.7581\\
    \midrule
    No Info Leak Map PT& \textbf{9.20} & 1.0343 & 0.7571\\
    W/ Info Leak Map PT & 9.84 & 1.0735 & 0.7614\\
  \bottomrule[1.2pt]
\end{tabular}
\vspace{-6pt}
\end{table}

In our unsupervised pre-training process, we adopt a method similar to MAE \cite{MaskedAutoencoders2021}, where the mask objects, including the unmasked first point, are \textit{not} passed to the encoder. While this eliminates the information leak from mask tokens, the encoder also loses the first point coordinates attached to masked objects. Therefore, we compare the prediction results with or without information leaks. In both scenarios, we ensure that the mask tokens pass through an equal number of transformer layers (i.e., one in our case) to exchange information. As shown in Tab. \ref{tab:unsupcompare}, we observe that the configuration without information leakage significantly outperforms its counterpart, affirming that the risk of information leakage outweighs the drawback of losing some positional data.

\section{Conclusion}
To expand and diversify the scant motion data for trajectory prediction, we propose a pipeline of synthesizing driving scenes and utilizing them via pre-training to provide general representations for initializing trajectory prediction models. In this study, we efficiently generate trajectories using the original map and its augmented variant. We also evaluate strategies for optimizing data utilization, finding that self-supervised pre-training performs better than the commonly-adopted alternatives. With our synthetic data generation and self-supervised pre-training pipeline, the fine-tuned prediction model outperforms the trajectory forecasting baseline by a large margin. Our research offers a comprehensive pipeline for data generation and utilization, providing a promising direction to alleviate the data scarcityor trajectory forecasting. 

\section{Acknowledgement}
Berkeley DeepDrive supports this work.







{\small
\bibliographystyle{IEEEtran}
\bibliography{egbib}
}

\end{document}